%% file: main.tex
\documentclass{sigchi-ext}
\usepackage[T1]{fontenc}
\usepackage{textcomp}
\usepackage[scaled=.92]{helvet} 
\usepackage{graphicx} 
\usepackage{balance}  
\usepackage{booktabs} 
\usepackage{ccicons}  
\usepackage{ragged2e} 
\usepackage{amsmath}
\usepackage{amssymb}
\usepackage{booktabs}
\usepackage{cleveref}
\usepackage{xcolor}
\usepackage{amsthm}
\usepackage{url}

\usepackage[utf8]{inputenc} 

\newcommand{\fig}[1]{Figure~\ref{#1}}

\newtheoremstyle{exampstyle}
  {0.7em} 
  {\topsep} 
  {} 
  {} 
  {\bfseries} 
  {.} 
  {.5em} 
  {} 
\theoremstyle{exampstyle} \newtheorem{challenge}{Challenge}
\newcommand{\RCref}[1]{\textbf{Challenge~\ref{#1}}}

\def\plaintitle{Counterfactual Explanations for Machine Learning: Challenges Revisited}

\def\plainauthor{Sahil Verma, John Dickerson, Keegan Hines}

\def\plainkeywords{XAI; Counterfactual Explanations; Trustworthy ML
  }

\title{\plaintitle}

\numberofauthors{3}

\author{%
  \alignauthor{%
    \textbf{Sahil Verma}\\
    \affaddr{University of Washington} \\
    \affaddr{Arthur AI} \\
    \email{vsahil@cs.washington.edu} 
    \email{sahil.verma@arthur.ai}
    }
    \vfil \alignauthor{%
    \textbf{John P.\ Dickerson}\\
    \affaddr{Arthur AI}\\
    \affaddr{University of Maryland}\\
    \email{john@arthur.ai} }
    
    \vfil \alignauthor{%
    \textbf{Keegan Hines}\\
    \affaddr{Arthur AI}\\
    \email{keegan@arthur.ai} \\
    }
    }

\definecolor{linkColor}{RGB}{6,125,233}
\hypersetup{%
  pdftitle={\plaintitle},
  pdfauthor={\plainauthor},
  pdfkeywords={\plainkeywords},
  bookmarksnumbered,
  pdfstartview={FitH},
  colorlinks,
  citecolor=black,
  filecolor=black,
  linkcolor=black,
  urlcolor=linkColor,
  breaklinks=true,
}

\begin{document}

\CopyrightYear{2021}
\setcopyright{rightsretained}
\conferenceinfo{CHI'21,}{May  8--9, 2021, Virtual}
\isbn{978-1-4503-6819-3/20/04}
\doi{https://doi.org/10.1145/3334480.XXXXXXX}
\copyrightinfo{\acmcopyright}

\maketitle

\RaggedRight{} 

\begin{abstract}
  \input{abstract}
\end{abstract}

\keywords{\plainkeywords}






\input{intro}

\input{desiderata}

\input{challenges}

\input{conclusion}


\balance{} 

\bibliographystyle{SIGCHI-Reference-Format}
\bibliography{sample}

\end{document}

%% file: abstract.tex
Counterfactual explanations (CFEs) are an emerging technique under the umbrella of interpretability of machine learning (ML) models. 
They provide ``what if'' feedback of the form ``if an input datapoint were $x'$ instead of $x$, then an ML model's output would be $y'$ instead of $y$.'' 
%
Counterfactual explainability for ML models has yet to see widespread adoption in industry.  In this short paper, we posit reasons for this slow uptake. 
Leveraging recent work outlining desirable properties of CFEs and our experience running the ML wing of a model monitoring startup, we identify outstanding obstacles hindering CFE deployment in industry. 
%

%% file: intro.tex
\section{Introduction}
\label{sec:intro}


Machine learning (ML) models are deployed broadly.  They now serve as components of morally-laden decisioning systems in hiring~\cite{hiring-ml}, criminal justice~\cite{parole-ml}, healthcare~\cite{medical-treatment-ml}, police patrolling~\cite{policing-ml}, and finance~\cite{credit-risk-ml}, among others. 
Usage in such critical applications necessitates trust---a multifaceted concept---in the underlying ML model.  
Core components of trust in ML models include provisioning, measurement, and monitoring~\cite{Mitchell19:Model} of fairness, privacy, and explainability---all in the context of the requirements and levels of comprehension of experts~\cite{Holstein19:Improving} and non-experts~\cite{Saha20:Measuring} alike.  In this work, we focus solely on a specific family of \emph{explainability} methods---counterfactual explanations---and identify barriers to their widespread deployment and use in industry.


\section{Counterfactual Explainability: A Tiny Primer}
Explainability research in ML either aims to develop inherently interpretable models~\cite{Rudin19:Stop} or to explain complex models. Examples of interpretable models include linear models, rule sets, shallow decision trees; examples of complex models include random forests and neural networks. 
Methods explaining complex models either take a holistic approach (global explanations) or explain individual predictions (local explanations). Global explanations generally approximate a complex model with an interpretable model; local explanations generally do not approximate the model. 


Counterfactual explanations (CFEs) are an emerging technique of local explainability.
They explain a prediction by calculating a change (usually minimal) in a datapoint that would cause the underlying ML model to classify it in a desired class. 
For example, if an individual were denied a loan request, a CFE might tell them that if they could increase their savings by \$5000, then their request would be approved. 
CFEs do not necessarily approximate the ML model and are therefore fidelitous to it. 
Unlike other explainability methods, CFEs aim to provide a precise and actionable recommendation to achieve a desired outcome. 


CFE in the context of ML was introduced by Wachter et al.~\cite{wachter_counterfactual_2017} in 2017.  (CFEs have existed in philosophy~\cite{Lewis1973:phil2} and psychology~\cite{Byrne:psycho1} from decades.) 
A burgeoning ML literature addresses different parts of the CFEs' desiderata, yet several challenges remain to be addressed before CFEs are widely adopted in industry. 
Counterfactual explainability is defined variously in the ML literature~\cite{verma2020CFsurvey,karimi2020survey}. 
Some works~\cite{karimi-imperfect:2020,karimi_model-agnostic_2020} differentiate between \emph{contrastive} and \emph{counterfactual} explanations--largely based on the presence of and assumptions about the causal relations between features--while others do not~\cite{wachter_counterfactual_2017,dhurandhar_explanations_2018}. 
Yet, all CFEs--contrastive and counterfactual alike--share a high-level goal of \emph{communicating} to stakeholders the underlying behavior of an ML model. 
Our present paper is largely agnostic to the detailed definition of CFEs;
indeed, the obstacles we identify hold broadly, and interfere with industry deployment of even the ``weaker'' notions of CFEs. 
Our goal is to surface challenges and directions for HCI and AI/ML researchers.



%% file: desiderata.tex
\section{Desiderata of Counterfactual Explainability}
\label{sec:desiderata}


\begin{figure}[h]
    \centering
    \includegraphics[width=0.9\columnwidth]{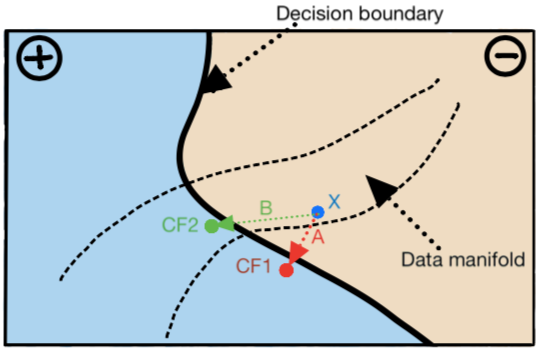}
    \caption{Two possible paths for a datapoint (shown in \textcolor{blue}{blue}), originally classified in the negative class, to cross the decision boundary. The end points of both the paths (shown in \textcolor{red}{red} and \textcolor{green}{green}) are valid counterfactuals for the original point. Figure adapted from Verma et al.~\protect\cite{verma2020CFsurvey}. }
    
    \label{fig:cf_fig}
\end{figure}


In this section, we list the desirable properties of CFEs as pointed out by recent surveys~\cite{verma2020CFsurvey,karimi2020survey}. These are, largely, the focus of the AI/ML sub-community focused on CFEs; their enumeration here enables discussion of some challenges we identify in the subsequent section.
\begin{itemize}[leftmargin=*]
    \item \emph{Actionability:} CFEs should consider the mutability and actionability of features when suggesting changes to them, for e.g. a CFE should not change immutable features like one's birth place or race. 
    
    \item \emph{Sparsity:} Instead of suggesting small changes to many features, a CFE is more amenable to action if it suggests changes to a few features, even if the magnitude of those individual changes are larger. 
    
    \item \emph{Adherence to data manifold:} Prior research has suggested a more actionable CFE to be closer to the manifold of the data used to train the underlying ML model. The training dataset represents the distribution of its features, and therefore a CFE far away from the dataset is likely unactionable. In \fig{fig:cf_fig}, the \textcolor{green}{green} CFE is closer to the data manifold than the \textcolor{red}{red} CFE. 
    
    \item \emph{Respect for causal relations:} In order to reflect real-world relations, a CFE should adhere to causal relations between features, for e.g. it should not suggest a decrease in one's age and degrees. 
\end{itemize}
\vspace{-1em}
The following are the desirable properties of the CFE generating algorithm.
\vspace{-1em}
\begin{itemize}[resume,leftmargin=*]
    \item \emph{Black-box and model agnostic CFEs:} An algorithm which can generate CFE for black-box models and is model-agnostic can be applied in various scenarios, including the case when the ML model is proprietary. 
    
    \item \emph{Fast CFEs:} An algorithm which can generate CFEs for multiple datapoints (from the same distribution) after a single optimization is faster and easy to deploy. 
\end{itemize}


%% file: challenges.tex
\section{Challenges in Operationalizing CFEs}
\label{sec:challenges}

Research has taken strides toward addressing different desired components of CFEs; however, much remains to be done before CFEs see widespread adoption in practice. In this section, drawing on intuition from our recent survey~\cite{verma2020CFsurvey} and our experience in industry, we enumerate some remaining challenges at the intersection AI/ML and HCI research, as well as policy and governance.

\begin{challenge}\label{ch:causal-lack1}
Lack of a structural causal model (SCM).
\end{challenge}
An SCM gives constraints on one feature (unary) or between multiple features (n-ary) of the dataset in the form of equations, and is therefore dataset specific. Most datasets 
do not have a readily-available SCM, making it difficult to generate CFEs that are actionable by individuals. A complete SCM is almost infeasible to have~\cite{causality:Pearl}, but even partial SCMs are sufficient for generating meaningful CFEs~\cite{mahajan_preserving_2020}.

\begin{challenge}\label{ch:causal-lack2}
Lack of interventional data. 
\end{challenge}
Most of the current datasets used in research and industry are observational, and therefore learning the complete causal graph is impossible~\cite{causality:Pearl,r60lecture,Peters2017}. Several causal discovery techniques exist that can learn causal graphs from interventional data~\cite{causal-discovery1}, e.g., data that shows how features of specific individuals evolved over time. Availability of such datasets can be useful in learning the causal relations and generating actionable CFEs. 


\begin{challenge}\label{ch:model-update}
ML models are not static (from~\cite{verma2020CFsurvey}). 
\end{challenge}
Most CFE generation approaches assume that the underlying ML model is stationary. However, this is not the case in many real-world settings; e.g., credit card companies and banks update their ML models frequently~\cite{bank-models}, and this must be taken into account when providing CFE to an individual.

\begin{challenge}\label{ch:interactive}
CFEs should be an interactive service, while guarding against privacy attacks (from~\cite{verma2020CFsurvey}). 
\end{challenge}
CFE should be provided as an interactive service. This provides an updated suggestion to an individual who could not precisely follow the prior advice. This however, could lead to privacy attacks causing leak of the data used to train the models, as it has been shown that with lot of queries to a ML model, one can infer information about the data used to train it~\cite{privacy1,privacy2,privacy3}. The CFE generating methods will need to guard against such concerns. 

\begin{challenge}\label{ch:notchange}
CFEs should tell what should not change (from~\cite{verma2020CFsurvey}). 
\end{challenge}
Along with what should change, a CFE should tell what should \emph{not} change. 
Consider a ML model used for loan prediction which takes \textit{income} and \textit{years in job} as input.  
The ML model rejected the loan request of an individual and suggested an increase in \textit{income}. 
Consequently, the individual changed their job and the feature \textit{years in job} was reset to 0. 
The model still rejected the loan request even after increase in \textit{income} as it did not specify that the other feature should not change~\cite{hidden_assumptions}. 
This situation could be averted if the individual could preemtively convey the proposed action and receive feedback (Challenge \ref{ch:interactive}).

\begin{challenge}\label{ch:bias}
Consider bias in the ML model (from~\cite{verma2020CFsurvey}). 
\end{challenge}
Most current CFE generation methods do not consider the potential bias in the ML model, which could lead to wide difference in difficult of attain a couterfactual state for different demographic groups~\cite{Ustun19:Actionable}. Research in fairness in ML has developed metrics of bias~\cite{verma_fairness} and methods to counteract it~\cite{dunkelau_fairness-aware}. CFE research should integrate method for generating CFE with the model bias consideration.

\begin{challenge}\label{ch:personal-pref}
Capturing personal preferences (from~\cite{verma2020CFsurvey}). 
\end{challenge}
Along with considering the actionability of features in global sense, CFEs should also consider their actionability in a local sense, i.e., personal preferences of an individual~\cite{mahajan_preserving_2020}. Most current CFE generation methods ignore this aspect.
Many industry applications further complicate this challenge by virtue of necessitating not just \emph{elicitation} of personal preferences, but also \emph{aggregation} of preferences across a diverse class of stakeholders. 

\begin{challenge}\label{ch:common-sense}
Generate common-sensical CFEs. 
\end{challenge}
The suggestions made by CFEs should align with common-sensical knowledge and human intuition, for e.g. it should suggest one to decrease their \textit{income} or \textit{degrees} in order to get a loan. The goal of generating robust CFEs will be beneficial to this challenge.

\begin{challenge}\label{ch:visual}
Lack of visualization of CFEs (from~\cite{verma2020CFsurvey}). 
\end{challenge}
Visualization can influence user behavior~\cite{Correll19:Ethical}, and since CFEs will be directly served to consumers with varying levels of technical knowledge, we need better ways to visualize CFEs.

\begin{challenge}\label{ch:regression}
Lack of understanding of how to apply CFEs in the regression (as opposed to classification) case. 
\end{challenge}
CFEs are deemed to be a method of explanation for ML models which classify their input into discrete classes. It would be beneficial to define CFEs for regression and develop methods that can generate them. 

\begin{challenge}\label{ch:vocab}
Unification of vocabulary and terminology within the CFE community. 
\end{challenge}
Currently, the CFE community uses varied terms such as ``contrastive,'' ``actionable recourse,'' ``algorithmic recourse,'' and others to describe similar ideas in the broad space of finding minimal and feasible changes to the features of an input datapoint in order to achieve some desired classification, perhaps subject to exogeneous constraints, stochasticity, and/or other forms of internal or external uncertainty. 
Recent articles have attempted to distinguish between, e.g., counterfactual and contrastive explanations~\cite{cfe-vs-cont,stepin2021survey,bodria2021benchmarking,karimi_algorithmic_2020,verma2020CFsurvey} by way of broad literature review.  Yet, those works also surface contradictory definitions and uses of some terminology by that literature. There is potential, and indeed need, to arrive at a consistent terminology to describe \emph{exactly} different manifestations of the broad idea of this form of explainability. 
Consistent terminology will help researchers to specifiy and address relevant problems, stakeholders to arrive at the same concept when discussing explainability techniques, and regulatory bodies to do their jobs accurately and appropriately.


\begin{challenge}\label{ch:regulatory}
Acceptance by regulatory bodies (from~\cite{verma2020CFsurvey}). 
\end{challenge}
The need for explainable methods in ML is driven in part by legal requirements, and the development of new explainability methods necessitates the creation of new policy. 
Our final challenge to the community is to curate and embrace the feedback loop between practitioners, researchers, and policymakers. 


%% file: conclusion.tex
\section{Conclusions}
\label{sec:conclude}

In this work, we have outlined a few important and challenging problems which need to addressed in order to promote wide acceptance of CFEs in AI/ML communities and consumer-facing industries. 
Among the set of challenges, \RCref{ch:interactive} and \RCref{ch:visual} are perhaps more direct to address for computer science researchers. 
Learning or deciphering causal models is known to be difficult and, often, impossible (\RCref{ch:causal-lack1} and \RCref{ch:causal-lack2}); indeed, understanding the tradeoffs between .  Capturing dynamism (\RCref{ch:model-update}), pre-defined notions of bias (\RCref{ch:bias}), baking-in common sensical knowledge (\RCref{ch:common-sense}), and encapsulating personal preferences (\RCref{ch:personal-pref}) in the CFE generating process are broad research directions with deep connections to computational social science and economics, operations research, and applied mathematics.  
\RCref{ch:vocab} and \RCref{ch:regulatory} will necessarily involve dialogue with a diverse set of stakeholders, perhaps moreso than the other challenges we have identified.
While each challenge brings important aspects to the utility of CFE methods, we believe that addressing these areas will increase the uptake of CFEs in practical application. These challenges should serve as promising future research directions.